# That was not what I was aiming at! Differentiating human intent and outcome in a physically dynamic throwing task

Vidullan Surendran · Alan R. Wagner

**Abstract** Recognising intent in collaborative human robot tasks can improve team performance and human perception of robots. Intent can differ from the observed outcome in the presence of mistakes which are likely in physically dynamic tasks. We created a dataset of 1227 throws of a ball at a target from 10 participants and observed that 47% of throws were mistakes with 16% completely missing the target. Our research leverages facial images capturing the person's reaction to the outcome of a throw to predict when the resulting throw is a mistake and then we determine the actual intent of the throw. The approach we propose for outcome prediction performs 38% better than the two-stream architecture used previously for this task on front-on videos. In addition, we propose a 1D-CNN model which is used in conjunction with priors learned from the frequency of mistakes to provide an end-to-end pipeline for outcome and intent recognition in this throwing task.

**Keywords** intent recognition · human robot interaction · surface cues · computer vision

## 1 Introduction

Robots designed to assist humans should be capable of dynamically varying their behaviour in response to the human's needs. Prediction of intent allows for efficient task planning by reasoning about what is being done

Vidullan Surendran
301C Engineering Unit C, Pennsylvania State University,
University Park, USA 16801
E-mail: vus133@psu.edu

Alan R. Wagner
234B Hammond Building, Pennsylvania State University,
University Park, USA 16801
E-mail: azw78@psu.edu

and why (Sukthankar et al., 2014). Liu et al. showed that, in a collaborative scenario where humans are unable to explicitly communicate their intent, inference of human partner goals improved perceived and objective performance of a human-robot team, and that participants preferred to work with a robot that adapted to their actions (Liu et al., 2018). In addition, understanding intent in the context of a task informs us of deviations from normal operation allowing us to spot mistakes early. In a complex task with many steps, this could help prevent failures by bringing such deviations to the attention of the person and solutions could be provided to mitigate future problems.

Here, intent is defined as an internal state vector of the human subject that is indirectly observable. We use the terms intent and goal interchangeably in this text. Actions, on the other hand, are the external, observable, manifestations of intent. We consider an example to differentiate between intents and actions. Consider, for example, a pick and place task where participants are asked to manipulate objects based on criteria such as colour, shape, or object type (Jois and Wagner, 2021; Lin and Chiang, 2015; Lamb et al., 2017; Alikhani et al., 2020). An intent or goal could be to complete the task in a specified amount of time. To achieve this intent and perform the task, the person must pickup objects, decide if they fit the imposed criteria, and move them to the appropriate locations within a set amount of time. This leads to sub-goals such as 'pickup an object', 'determine what criteria it satisfies', and 'move it to the correct location'. In other words, intents are naturally hierarchical, contextual, and task dependent. Thus recognizing intent requires knowledge of the task.

On the other hand, actions are the observable physical interactions performed by the person to achieve these intents. Action recognition is typically concerned with identifying the physical motion performed by the person, for example, classifying the motion as a grasp, lift, move, or drop. Although action recognition is a key component necessary for recognizing intent, action recognition does not predict the intent unless there is a one-to-one correspondence between the intent and action. Mistakes preclude a one-to-one correspondence between the outcome of an action and the intent.

Humans are rarely perfect when executing complex real world tasks. Mistakes are common when the task is physical and dynamic. Mistakes can be attributed to incomplete information, misunderstandings, physical inability of the body to repeatedly and accurately replicate motion patterns, or a myriad of other factors. Little research has considered the challenge of recognizing intent during mistake riddled, physically dynamic tasks. For such tasks, relying on action recognition to determine intent ignores the possibility that the observed action is incongruent with the underlying intent. Yet, a person is usually aware of the errors they commit if they are able to observe the outcome of their action. We postulate that this awareness is expressed as observable surface cues such as body motion and facial expressions generated by the person. We then hypothesize that using surface cues will allow us to detect congruity between the subject's intent and the outcome of their action. Used in conjunction with a known task model, this allows us to predict intent, or at the very least reduce the search space of possible intents.

This work focuses on predicting intent in the physically dynamic task of throwing in the presence of mistakes. When mistakes occur, the intent of a human thrower is different from the target the throw hits. The methods proposed for human intent recognition in a throwing task (Zhang et al., 2019; Li et al., 2018, 2020) omit throws that are mistakes and rely on predicting the outcome of the throw which is considered to be the subject's intent. We argue that this is unrealistic as the inherent difficulty of a throwing task leads to a high percentage of mistakes. To this end, we contribute:

- A dataset consisting of 1227 throws generated by 10 different participants captured from 3 different viewpoints which includes mistakes [1]
- Prediction of the outcome of a throw when the trajectory of the ball is only partially visible
- Recognition of the presence of mistakes relying on the human's reaction to the outcome
- An end-to-end pipeline for intent prediction accounting for the possibility of mistakes
- Measures of human performance on this task which can be used as a baseline comparison

---

[1] Request access at https://sites.psu.edu/real/datasets/

## 2 Related Work

Identifying key frames in the throwing motion and intent recognition relies on determination of human body pose. This can be achieved with a motion capture system or estimated from RGB-D (RGB with an added Depth channel) images. Image based methods have reduced accuracy, but are not restricted to operate within a tracked volume of space. They also forego the need for high performance sensors making them advantageous in robotic applications. Alpha Pose (Fang et al., 2017) for instance, uses a spatial transformer architecture to generate two dimensional human pose from RGB images and achieves 72.3 mean average precision (mAP) on the COCO dataset and 82.1 mAP on the MPII dataset. A commonly used closed source alternative is the Microsoft Kinect V2 which uses RGB and depth from its IR camera to estimate pose (Shotton et al., 2011).

Huang and Mutlu tracked human gaze to develop an anticipatory control method enabling robots to proactively perform tasks based on predicted intent (Huang and Mutlu, 2016). Gaze patterns from 276 episodes were used to train a support vector machine (SVM) classifier based on four features - number of glances, duration of first glance, total duration, and most recently glanced item. They report that their model predicts user intent approximately 1.8 seconds prior to human verbal request with an accuracy of 76%. Yang et al. proposed a task-based method that captured hand gestures and eye movement to recognize user intent, arguing that hand and eye movement represents a vast majority of nonverbal expressions of intent (Yang et al., 2016). Surendran and Wagner showed that surface cues in the form of upper body gestures could be used to determine if a human was trustworthy or not in the context of a card playing game involving deception (Surendran and Wagner, 2019). Their approach achieved prediction rates that were on par with humans performing the same task. More recently, they extended this work to use facial emotions as features for end-to-end prediction of deceptive intent (Surendran and Wagner, 2021).

Yu and Lee used multiple timescale recurrent neural networks to tackle the problem of intent recognition from recognised motions (Yu and Lee, 2015). Their experiment defined eight types of motion primitives and five intents that were related to drinking, moving, and pouring from a cup or a container. One of the strengths of this work is that they acknowledge that a time sequence of action primitives has to be considered to determine intent as many intents may share the same subsequence of actions. A limitation of their approach is that action sequences that map to an intent have to be pre-defined and known. This was acceptable for the tasks they explored, but is not reasonable for the throwing task where the sequence of primitives that are exhibited is unknown and varies between subjects.

Classification based on human cues generated from videos can be formulated as a time series classification task and can be tackled using algorithms such as nearest neighbour (Lee et al., 2012), dynamic time warping (Jeong et al., 2011), LSTM architectures (Karim et al., 2017) and recently, 3D convolutional networks (CNNs) and temporal CNNs suited for multi-channel data such as images (Akilan et al., 2019). A study with over three million automatically sampled experiments on 75 time series data-sets showed that single metric classifiers very rarely outperform dynamic time warping (Lines and Bagnall, 2015). This provides an empirical argument for ensembles models such as random forests and multi-branch neural networks.

In a task using overarm throws Maselli et al. found that for most participants it was possible to predict the region where the ball would hit a target as early as 400-500 milliseconds before release (Maselli et al., 2017). They were also able to predict which one of 4 targets were hit with an accuracy of 80% and that towards the end of the throwing action, the throwing arm motion provides the most information. This study was only concerned with where the ball struck the target and not where the participant intended to throw it. It effectively ignores mistakes and the intent of the subject but provides evidence that if extremely accurate joint predictions are available, motion alone can predict the outcome of a throwing task. This method, which used a 16 camera motion capture system to generate highly accurate joint positions, is not ideal for robotic applications in the wild.

Li et al. utilized a Kinect V2 stereo camera to predict the outcome of a throwing task with 9 targets (Li et al., 2020). The human pose estimates generated by the Kinect were used to determine the start of the throwing action. A single camera positioned head-on to the subject, similar to the setup shown in Figure 1 was used to collect data. They also compared various pre-processing schemes such as K-random sampling of the input frames, using the entire clip, and a combination of both of these to overcome the problem of having frames in the video that were not relevant to the task. The model used for prediction of the throwing outcome was based on the two-stream architecture (Simonyan and Zisserman, 2014) which uses optical flow in conjunction with RGB images.

A majority of the published works that used a throwing task for intent recognition (Zhang et al., 2019; Li et al., 2018, 2020) excluded trials where the participant did not strike the target they were instructed to hit, i.e. mistakes. Excluding mistakes results in artificially mapping the thrower's intent to the realized outcome. We argue that this exclusion reduces the real-world validity of the proposed intent recognition methods since in this task a significant percentage of throws result in mistakes. During our data collection we found that 47% of throws were mistakes and this is detailed in Section 3.2. Moreover, our attempts to reproduce the reported performance from (Li et al., 2020) were not successful and this is explored in Section 7.

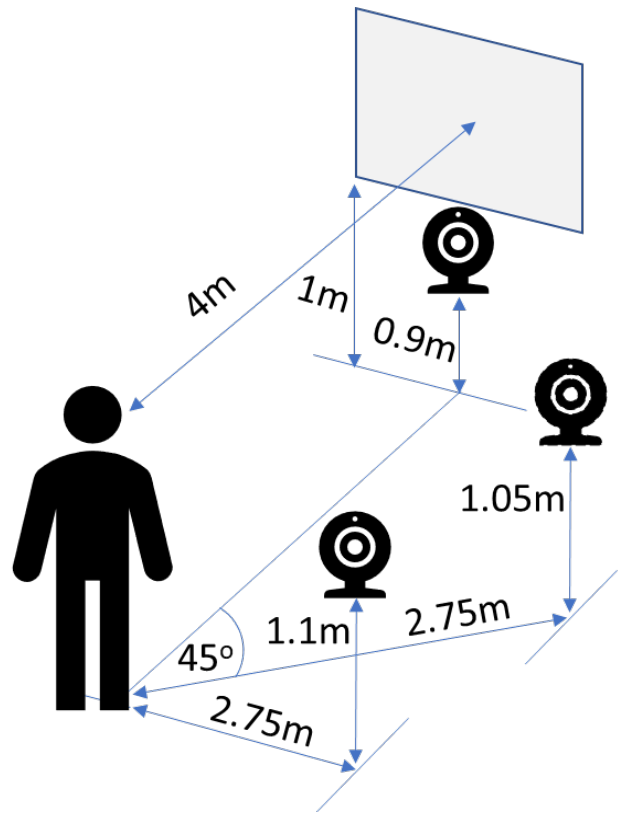

Fig. 1: Throwing task setup with relevant dimensions

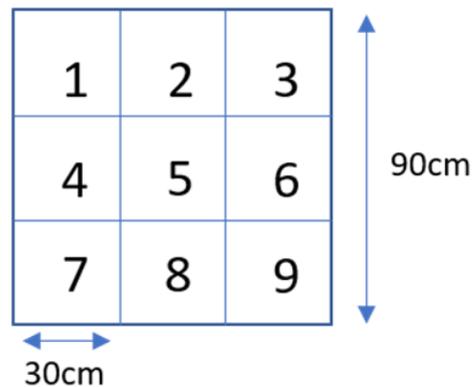

Fig. 2: 3x3 target grid with dimensions and zone labels

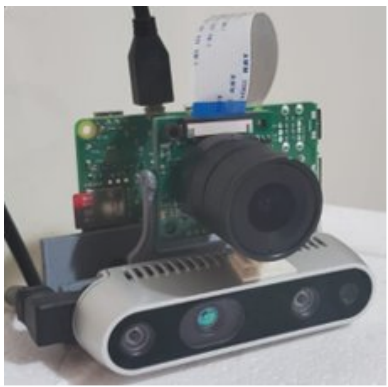

Fig. 3: Dual camera setup consisting of the Intel D435 and a Pi camera fitted with a 50mm lens. The Raspberry Pi 4 used to process images from the Pi Camera can be seen vertically mounted behind the cameras.

## 3 Throwing Task

We seek to study human intent recognition during physically dynamic tasks. We characterize physically dynamic tasks as tasks that require a rapid series of physical motions, tend to result in improved performance with training, and as a consequence, tend to result in a high rate of mistakes unless the person is highly trained. A throwing task is one example of a physically dynamic task. Throwing tasks tend to result in a high number of unforced errors unless the subject has extensive previous training. Our version extended the throwing task described in (Li et al., 2018) to include more participants, more camera angles, and our dataset included samples where the thrower made a mistake.

### 3.1 Experimental Setup

The overall setup is shown in Figure 1 which is annotated with the relevant dimensions. Three camera units placed at 0°, 45°, and 90° with respect to the participant allowed the capture of front-on, angled, and side-on images of a right handed thrower encompassing most of the commonly encountered camera viewpoints. A distance of 2.75 meters was used for the 45° and 90° camera units due to space restrictions, but this still allowed us to fit a 6ft tall subject with their arms extended vertically overhead in the camera frame. The 0° camera unit placed under the target 4 meters away replicates the setup used in related work (Surendran and Wagner, 2021; Li et al., 2020, 2018).

The target was a 3x3 grid with nine equally sized zones numbered as shown in Figure 2. Each of the camera units consisted of an Intel RealSense D435 and a Raspberry Pi camera mounted as shown in Figure 3.

The D435 cameras were mainly used to capture wide angle RGB images of the subject at a resolution of 848x480 pixels and a frame rate of 30FPS. These camera were auto focused at infinity. The Raspberry Pi cameras, fitted with a 50mm lens, were manually focused such that the participant's upper body and face was captured at a higher resolution of 1080p and a frame rate of 30FPS. Sample images from the cameras can be seen in Figure 5 (right).

The internal clocks of the three RealSense cameras were hardware synchronized to simultaneously trigger the frame capture. The Raspberry Pi cameras do not support this functionality and thus the software reported timestamp from each captured frame was used to approximately match them. To ensure the clocks of the three Raspberry Pis connected to the cameras had a common reference, we synchronized them to a local Network Time Protocol (NTP) server. Using the six timestamps (three of which are identical due to the hardware synced D435s), we are able to associate each RealSense frame to a corresponding frame from the Pi Cameras. As all cameras were run at 30FPS, at worst we expect to have a frame capture time skew of 33ms between the Pi camera image and the D435 image.

### 3.2 Data Collection

IRB approval was obtained and 10 right-handed subjects were each compensated $15 USD to perform the throwing task for an hour. The subject pool consisted of 6 men and 4 women who were physically able, aged 25-35, with a recreational level of throwing proficiency. The participants ranged in height from 4ft 11in to 6ft 1in approximately. Subjects were asked to sign a consent form permitting the recording and sharing of the experiment before they began. The consent form explained the task, but did not mention anything related to intent recognition, mistakes, or their facial expressions and body movement being used for predictions. This was to minimize over-exaggerated reactions they might unconsciously perform in order to aid the study.

Subjects were instructed to place a foot in the vicinity of an 'X' mark affixed to the floor so as to stay within view of the cameras. They were told that they could use any overhand throwing motion to throw the ball. A reference of the target numbering was affixed above the target to assist the subject in case they forgot what each target zone was called. Each throwing trial consisted of the experimenter starting video collection and calling out the target zone the participant was to aim at. Once the subject threw the ball, the experimenter would stop video collection a couple seconds after the ball hit the target. In certain cases, where the subject was still noticeably reacting to the outcome of their throw, video collection was continued until a subjective assessment was made to determine when to stop video collection. After the recording was stopped, a generic buzzer sound was played to notify the subject that they were free to move away from the throwing location to pick up a ball for the next throw. Subjects were instructed to only take one ball at a time. Nine identical balls were used so that the subject could throw a ball at each one of the nine target zones before having to collect them. The experimenter sat behind a screen obscured from the subject's field of view to minimize visual interaction with the subject. Even then, subjects had a tendency to look towards the general direction of the experimenter after completion of a throw.

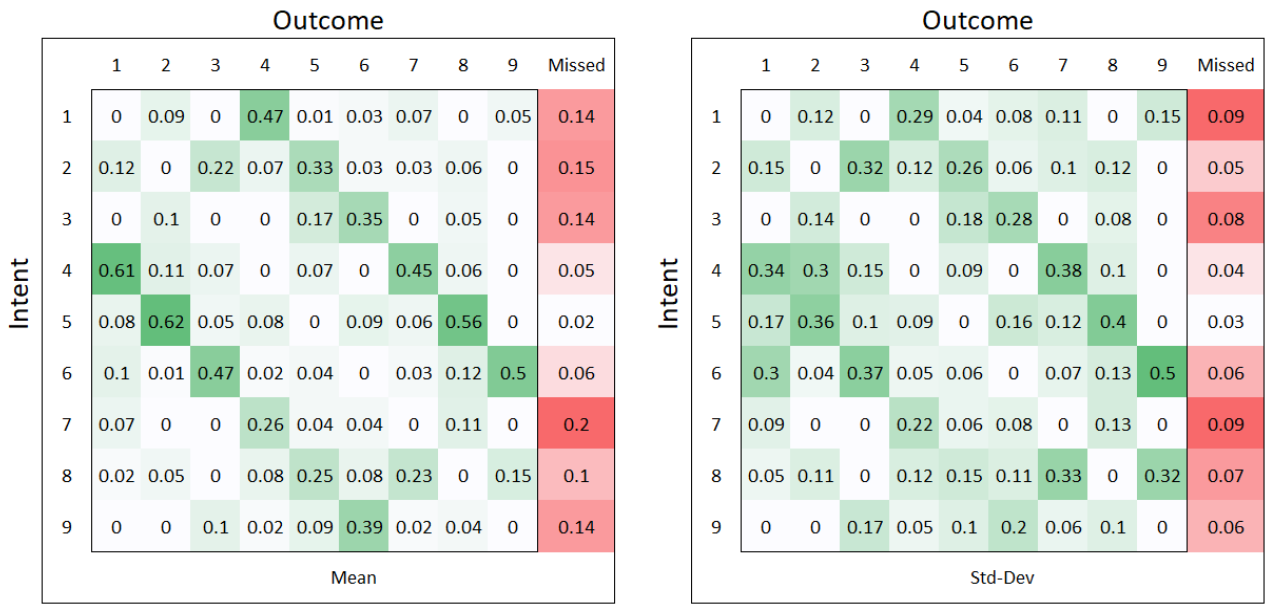

Fig. 4: Probability of *intent* target given observed *outcome* target in the presence of a mistake, i.e. the subject misses the target they aimed at. 'Missed' refers to the subject missing the target grid entirely, i.e. did not hit any of the 9 zones.

The target aimed at was considered to represent the subject's *intent*, whereas the target the ball actually struck was considered to be the *outcome*. A round was defined as all 9 target zones being selected as the *intent* exactly once. The sequence of *intent* targets was randomly generated each round to minimize learning effects and mitigate extraneous factors such as lighting variations from biasing the data. For example, if one target zone was aimed at multiple times before moving on and the lighting changed, one could differentiate between the targets solely based on image brightness. Learning cannot be completely eliminated as participants tend to improve with practice. Accuracy improvement with practice was counteracted by the decline in performance caused by fatigue over the course of the hour but the size of these effects is unknown.

### 3.2.1 Ethical Considerations

While collection of such data is predicated on obtaining IRB approval which also governs how the data is shared, the researchers involved who are intimately familiar with the information should carefully consider the privacy and equity issues inherent in its distribution; especially for datasets such as ours containing non-anonymized videos of the participants. With the advent of big data and psychology research conducted via mining social media data, there is an ongoing discussion in the community on best practices to protect the subjects when using publicly available video and social media data (Kraut et al., 2004; Legewie and Nassauer, 2018; Derry et al., 2010; Townsend and Wallace, 2017; Zimmer, 2017).

The subjects that participated in our research study are given the authority to revoke consent at any point which would result in the deletion of their data. But, once distributed, it is impossible to guarantee that other users of the dataset have deleted local copies of the data. This consideration affected how we chose to share the data. We only make the dataset available through explicit request which includes collecting information pertaining to who the users will be, how they will use the data, and how they will protect the data to ensure the privacy of the participants. This allows us to track the users and inform them when participants withdraw consent. We realise that this relies on trusting the users granted access will follow the protocol and in practice limits access to researchers belonging to trusted institutions since they can reasonably be expected to behave ethically. While this is in not in the spirit of democratizing science, we consider the ability of our subjects to withdraw consent paramount. An alternative is drafting a legally binding contract for users of the dataset which reduces the emphasis on trust potentially allowing access to a larger user base, but this is complicated by the global nature of research and lies outside the scope of expertise of the authors.

Table 1: Per subject throws, mistakes, and accuracy

| Subject  | 1    | 2    | 3    | 4    | 5    |
|----------|------|------|------|------|------|
| Throws   | 135  | 200  | 63   | 117  | 82   |
| Mistakes | 78   | 56   | 38   | 56   | 61   |
| Accuracy | 0.42 | 0.72 | 0.40 | 0.52 | 0.26 |
| Subject  | 6    | 7    | 8    | 9    | 10   |
| Throws   | 99   | 135  | 135  | 153  | 108  |
| Mistakes | 69   | 63   | 35   | 64   | 62   |
| Accuracy | 0.30 | 0.53 | 0.74 | 0.58 | 0.43 |

Table 2: Mean (std-dev) probability of a mistake when aiming at each target zone (Round-off error results in a sum of 1.01).

| 0.11 (0.02) | 0.12 (0.03) | 0.12 (0.03) |
|-------------|-------------|-------------|
| 0.11 (0.02) | 0.11 (0.03) | 0.10 (0.04) |
| 0.12 (0.02) | 0.11 (0.03) | 0.11 (0.03) |

Table 3: Number of times each target was hit, i.e. was the *outcome* target. 192 throws completely missed the target

| 1   | 2   | 3  | 4   | 5   | 6   | 7  | 8   | 9  |
|-----|-----|----|-----|-----|-----|----|-----|----|
| 115 | 107 | 95 | 129 | 150 | 141 | 97 | 115 | 86 |

As practitioners implementing data driven models and generating datasets that might become a benchmark, we must consider data biases that can arise due to subject demographics. Since the majority of HRI research arises out of well funded labs in a few countries, there is an under-representation of certain races, genders, and cultures, leading to computational models trained on such data being discriminatory (Garcia, 2016; Cheuk, 2021; Zou and Schiebinger, 2018). This is a very difficult problem to overcome as recruiting numerous diverse subjects is extremely challenging and is something to be cognizant of when deploying models that are trained on our dataset.

### 3.3 Statistics

In total, 1227 throws (mean = 123, std-dev = 37) were made with 192 throws completely missing the target grid and 582 throws resulting in mistakes. Table 3 shows the number of times a target zone was hit regardless of it being selected as the *intent* target. While the number of mistakes varied based on the participant's skill level, the mean proportion of mistakes to valid throws over all participants was found to be 0.51 (std-dev = 0.15). Per subject counts are shown in Table 1. We can discount any significant bias towards a particular target zone since the probability of making a mistake was found to be independent of the participant's *intent* target as shown in Table 2.

Using the frequency counts of the *outcome* target given a particular *intent* target, we calculate a prior probability of the *intent* target given an observed *outcome*. Figure 4 shows this matrix of probabilities for the aggregate data of all participants in the presence of a mistake. The diagonal elements are 0 since otherwise it would not be an erroneous throw. This representation allows us to infer that certain targets are highly unlikely to be the *intent* given a particular *outcome*, for example, zones 7 or 8 being the *intent* when the *outcome* was target zone 3. This makes intuitive sense as one is unlikely to hit the bottom left target zones when aiming at the top right target zone.

## 4 Computational Pipeline

Given a video of a subject throwing a ball, we would like to predict the *intent* target. This requires us to determine if the throw is erroneous, which in turn requires the determination of the *outcome* from videos which only partially capture the trajectory of the ball.

Our approach uses RGB image data to (a) determine the target zone that will be struck, i.e. predict the *outcome* target (discussed in Section 4.3), and (b) predict the target that was being aimed at, i.e. the *intent* target (discussed in Section 4.4). We begin by determining the video frame that signals the end of the acceleration phase of the throwing action, referred to as the *throw frame*. This information is used to split the video into the *throwing* segment and the *reaction* segment. Assuming a subject's reaction to the outcome would only be observable after they throw the ball, we can reduce the number of uninformative frames by using the *throwing* segment from the D435 camera which captures the entire subject to address problem (a), and the *reaction* segment from the Pi camera which provides higher resolution images of the subject's upper body to address problem (b).

### 4.1 Generation of 2D Pose

Human pose was predicted using Alphapose which is a deep learning based multi-person pose estimator (Fang et al., 2017). It outputs 26 joint key-points using the COCO dataset format which includes the location of the following joints: nose, eyes, ears, shoulders, elbow, wrist, hips, knees, ankles. Missing joints, resulting from one part of the body occluding another or low quality images, were linearly interpolated in the temporal direction. The 2D pixel pose estimates were not temporally stable as they are estimated on a frame-by-frame basis which varied slightly over the video even in frames with negligible movement. Moreover, interpolation can result in sudden jumps in joint position when there is excessive joint motion. A Butterworth filter with cutoff frequency of 2Hz was used to reduce high frequency noise over the temporal dimension due to its maximally flat response. The effect of filtering can be observed in Figure 5 (bottom left).

### 4.2 Determination of throw frame

For the 0° cameras, we estimate the the end of the acceleration phase of the throwing motion using the following heuristic:

- The magnitude of the velocity of the throwing arm wrist joint is expected to be sinusoidal going from a standstill (gather), to a minima (cocking), to a global maxima (acceleration), and then approach zero as the arm decelerates at the end of the action (deceleration/follow up).

The scoring metric used for the $i^{th}$ frame of a video containing $n$ frames was,

$$score_i = \frac{\|v_i\|_2 - max(V)}{max(V)},$$
$$V_i = \|v_i\|_2 \ , \ i = 0, 1, ..., n \quad (1)$$

where $v_i$ denotes the velocity of the throwing wrist joint in the $i^{th}$ frame, and $V$ is the vector containing the $L_2$ norm of velocity for the entire throwing sample. Since this score uses velocity estimated from noisy joint pixel data, we used a Savitzky-Golay filter to fit a $2^{nd}$ order polynomial using a convolution window length of 11 steps. The score along with the x and y pixel coordinates for a sample are shown in Figure 5 (top left).

From approximately frame 60 to 90 we observe the sinusoidal pattern that corresponds to the participant cocking their arm back, accelerating towards the target, and then slowing down during the follow through. At the end points, we tended to obtain spikes due to the filtering approach used. We thus ignore the first and last 10% of the frames when searching for the frame with the maximal score. In the example considered (Fig. 5), the score reached a peak at frame 77.

The 45° and 90° cameras are able to capture participant joint translation normal to the target grid plane, and thus we used the following heuristic:

- The distance between the wrist and the body would be at a maximum at the end of the acceleration phase due to maximal elbow extension.

This translates to finding the frame with the maximal signed offset between the x-pixel values of the throwing wrist joint and the hip joint. If only the wrist position is considered, there is a possibility of false positives whenever the subject moves backwards in the frame while throwing.

Figure 5 (right) confirms that the applied heuristics are able to determine the end of the acceleration phase of the throwing action. The hardware synced D435 images all show the same frame from different viewing angles as expected. The prediction for the Rasberry Pi cameras differed slightly as these software synced cameras cannot guarantee frame capture at the exact same instant.

Inspection of the predicted throw frames revealed that the two most common sources of inaccuracy in throwing frame prediction were blurry image frames, and taller participants moving too far forward while throwing resulting in their wrist moving outside the camera frame. Unfortunately, this is commonly found in the Rasberry Pi camera data due to the zoom lens reducing the field of view, whereas the D435 cameras with their wide angle lenses were less affected by subject translation. We did not manually correct the erroneous predictions because our system is meant to be end-to-end autonomous.

### 4.3 Outcome Prediction

Here we try to predict the *outcome* target the subject struck irrespective of their intent. Focusing solely on the joint pose results in poor performance in outcome prediction as a multitude of throwing actions can result in an identical ball trajectory. Two identical throwing motions with a minimally varying release position would result in different outcomes. Detection of release position from the 30FPS camera streams was not possible since the fast throwing motions resulted in either blurry frames or the exact moment of release not being captured. The outcome prediction model overcomes this limitation by tracking the position of the ball starting from its initial position coincident with the throwing wrist joint and its trajectory after release until it leaves the camera's field of view. This implicitly encodes the assumption that the wide range of throwing and follow through motions employed by the subjects is encoded by the position of the ball greatly reducing the dimensionality of our problem.

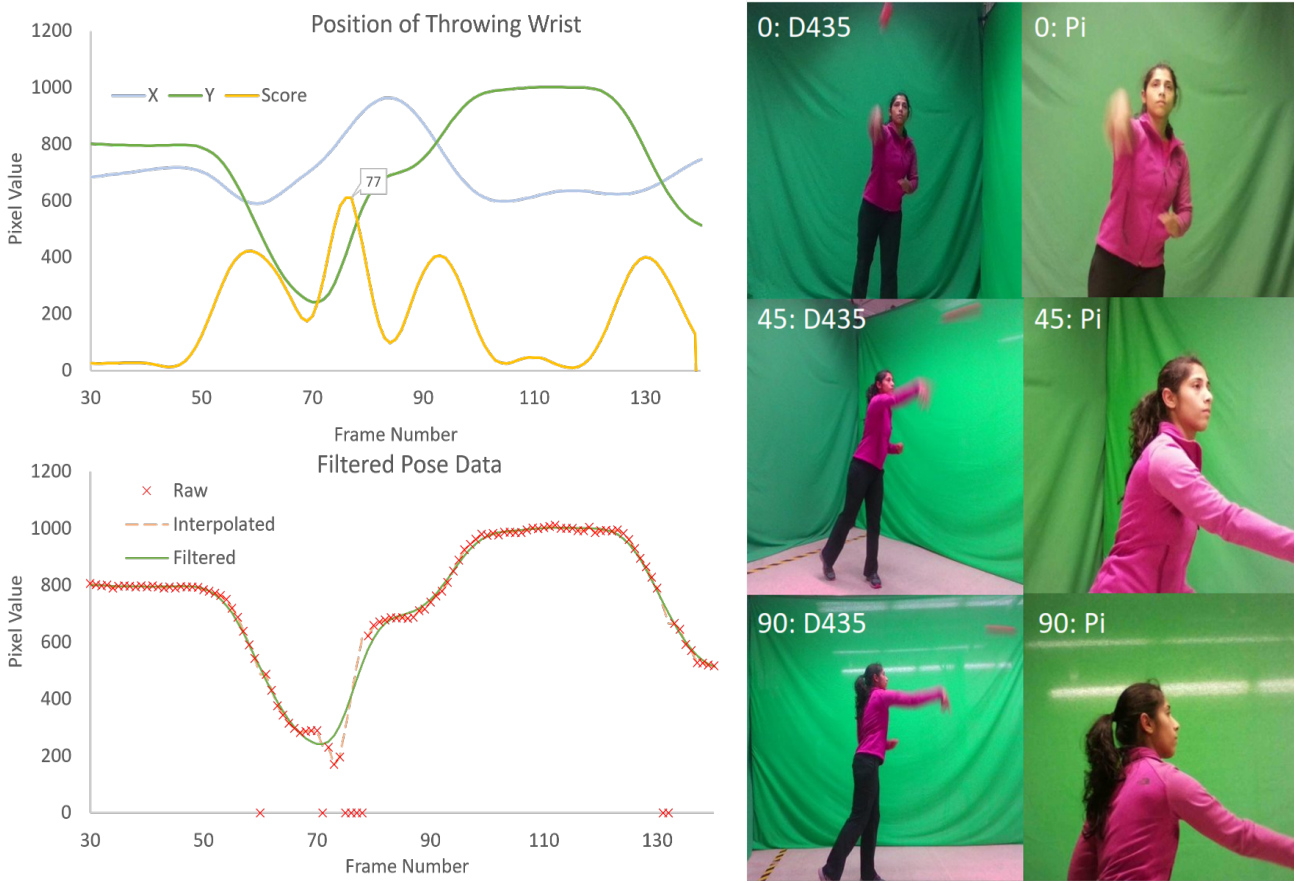

Fig. 5: **Top left** graph shows the filtered X,Y values of the throwing wrist along with the composite score for a sample captured by the 0° D435 camera. The score was scaled for graphing and the maximum value was observed at frame 77 denoting the throw frame. **Bottom left** shows the raw, interpolated, and filtered throwing wrist Y coordinate values illustrating the effect of preprocessing discussed in Section 4.1. **Right** shows the throw frame from all 6 cameras.

Table 4: Layer Parameters for 1D CNN Model. Key: 'c' is channels, 'k' is kernel size, 's' is stride, 'n' is number of units, 'act' is activation function. 'A', 'B' and 'C' refer to the model sections from Figure 6 (bottom).

|   | Input | Conv1D | Dropout | Conv1D | MaxPool |
|---|---|---|---|---|---|
| A | shape = (Batch, 30,7) | c=8, k=3, s=1 | 0.1 | c=16, k=3, s=1 | k=2 |
| B | shape = (Batch, 30,7) | c=8, k=9, s=1 | 0.1 | c=16, k=9, s=1 | k=2 |
|   | Concatenate | Dropout | Dense | Dense |   |
| C | output = (Batch, 320) | 0.1 | n = 20, act: relu | n = 2, act: softmax |   |

Due to the large amount of movement inherent in the throwing action, and the ball overlapping the subject in the 0° camera, dense optical flow could not be employed to track the position of the ball; The small moving object, that is the ball, was indistinguishable from background noise. We use the estimate of the throw frame number, $N_{tf}$ to add robustness to this pro-cess. The ball is assumed to overlap the throwing wrist joint position for the frames numbered 0 to $N_{tf} - 10$. Further, we assume that the ball is outside the field of view of the cameras for frames numbers greater than $N_{tf} + 5$. For the frames in between these ranges, we convert our colour space to HSV, and threshold the image to detect regions with the same colour as our ball. The HSV colour space makes our threshold operation robust to varying lighting conditions compared to the RGB colour space. We filter the resultant binary image using the morphological operations of closing with a 3x3 kernel. Contours are then detected and the center of a rectangular bounding box enclosed around the largest detected connected region of pixels is used as the location of the ball in that frame.

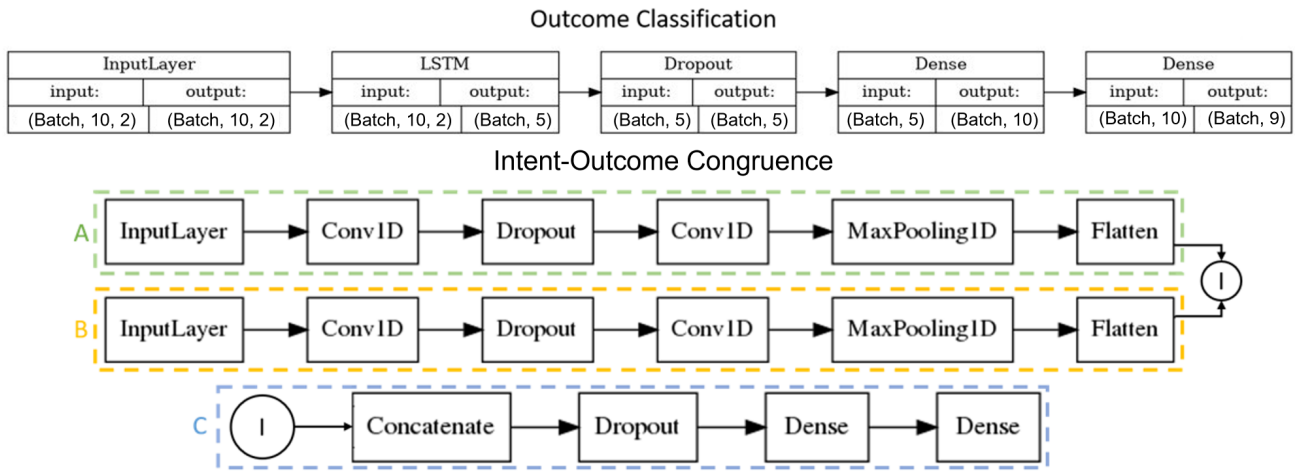

Fig. 6: **Top** shows the LSTM model used to classify 2D pose data into one of 9 outcome classes. 'Batch' refers to the variable data batch size used during training/inference. **Bottom** shows the Multi-branch 1D CNN model used to detect congruence between outcome and intent using features from a pre-trained emotion model. A and B denote two input branches whereas C is the concatenated branch. The layer parameters are shown in Table 4.

The *outcome* target is predicted by using a feature vector containing the positions of the ball in the $N_{tf} - 5$ to $N_{tf} + 5$ frames. The position of the ball implicitly encodes the position of the throwing wrist joint position in these frames and thus our feature vector is of length 2 (x, y pixel position in image). An LSTM model was used to classify the feature vector into one of 9 target classes and its architecture is shown in Figure 6 (top).

### 4.4 Intent Prediction

This is a two-step process where we first recognize if the outcome is congruent with the intent and subsequently predict the intent of the participant. Images from the Pi camera provide a clearer view of the upper body and face of the subject for the surface cues we hope to extract. One second worth of frames (30 frames) from the *reaction* segment beginning 10 frames after the identified throw frame is used. The first 10 frames after the release of the ball are ignored as these capture the subject in the follow through phase of the throwing action. This method implicitly assumes that the subject visibly expresses their internal state and the most relevant information is obtained within one second of the completion of the throwing action.

#### 4.4.1 Intent-Outcome Congruence

It is challenging to recognize or predict a subject's reaction without a prior model of their behaviour. Humans express themselves using a large number of primitive gestures and combination of gestures. The datasets used for full body gesture recognition such as NTU RGB+D (Shahroudy et al., 2016), NTU RGB+D 120 (Liu et al., 2020), A Large Scale RGB-D Dataset for Action Recognition (Zhang et al., 2018), two-person interaction dataset (Yun et al., 2012), and others all consider everyday activities and are segmented to include only a primary gesture per video sample. They do not consider reactionary body gestures which is required for determining the participant's reaction in our throwing task. Moreover, the problem of continuous gesture recognition and segmentation is an active area of research with problem specific knowledge used to segment gestures.

Using the full range of expressive gestures and human expression would be preferred, but given the lim-itations of current gesture recognition algorithms, we employed facial reactions. While limited, prediction of facial emotion based on front-on images is a well studied problem. Detection and cropping of faces in each frame was carried out using an implementation of FaceNet available through the OpenCV computer vision software library (Schroff et al., 2015). A CNN pre-trained on the FER 2013 dataset (Arriaga et al., 2017) which contains 35,685 images representing seven expressions, namely, happiness, neutral, sadness, anger, surprise, disgust, and fear was used to generate feature vectors from the images of the cropped faces. We modified the CNN by removing the output softmax classification layer and using the flattened weights from the last 2D global average pooling layer as our feature vector.

Our proposed solution relies on detecting the facial emotion response and using this to infer if the subject has committed a mistake (Surendran et al., 2021; Surendran and Wagner, 2021). By using a pre-trained facial expression model, we leverage priors learned from a much larger dataset and eliminate the need to label our throwing dataset. We employed a multi-branch 1D CNN model whose architecture is shown in Figure 6 (bottom) and layer parameters are shown in Table 4. The binary outcome of the model corresponds to the observed outcome fulfilling the subject's intent or the outcome being different from the intent indicating the presence of a mistake.

### 4.4.2 Prediction

The output of the intent-outcome congruence model informs how we predict the *intent* target. If the intent is congruent with the outcome, the throw did not result in a mistake and the *intent* target is predicted to be the same as the *outcome* target. Otherwise, we can rule out the observed outcome being the intent and are left with 8 possible target choices. We generated a-priori probabilities as discussed in Section 3.3 and use this distribution to predict the intent target. Given an *outcome* target, the target zone with the maximum probability is chosen as the *intent* target. If multiple *intent* targets have the same probability, then one of them is randomly selected.

While certain probabilities were calculated to be 0 as these events were not observed during data collection, this might not hold true during inference. Add-one smoothing was used to assign a minimal but non-zero probability to prevent the model from discounting target zones with a probability value of 0. We do not consider online updates of this probability matrix which would negate the need for this smoothing.

Table 5: Average human labeler performance on the tasks. Performance metric listed in parenthesis.

|  | 0° Cam | 45° Cam | 90° Cam |
|---|---|---|---|
| Outcome (Accuracy) | 0.43 | 0.23 | 0.29 |
| Congruence (F1/MCC) | 0.69/0.14 | 0.74/0.33 | 0.72/0.23 |
| End-To-End (Accuracy) | 0.33 | 0.16 | 0.25 |

## 5 Human Performance as a Comparison

Due to the limited literature exploring intent recognition on tasks that include mistakes, there is a dearth of models that can be used for comparison. We recruited two humans labelers to perform the tasks of outcome, intent-outcome congruence, and intent recognition. This provides another comparison baseline in addition to random chance.

The labelers were shown 50 randomly sampled throwing videos from the 1035 throws that hit the target grid to help them learn each participant's tendencies, referred to as the learning phase. Then their performance was evaluated on 50 new videos randomly sampled from the dataset, referred to as the testing phase. The labelers were allowed to replay the video as many times as they wanted. They were asked to predict the *outcome* target, if the participant hit the correct target (intent-outcome congruence), and then predict the *intent* target. This process was repeated for the 0°, 45°, and 90° cameras ensuring the labelers saw the same video samples for all three views. Since the labelers were not provided the ground truth after each testing phase, seeing the same samples should not increase performance as a result of memorization of the correct answer. The testing phase used the same samples for the three views to prevent sample selection from skewing the results for the different views.

Once the human labelers completed the task, an informal interview revealed the strategies they employed. They said that they tried to use the trajectory of the ball and its position in the image to determine the *outcome* target. They also considered how fast the participant moved to extrapolate the ball's trajectory once it left the frame. In samples where the ball immediately went out of frame, or when the participant's limbs were not visible in the frame (commonly encountered in the 45° and 90° views), they expressed that 'educated guesses' were the only option. For the 45° and 90° views, the labelers mentioned that they primarily focused on predicting the row of the target grid the ball would strike as it was extremely difficult to predict the column due to the lack of depth data.

Both human labelers felt that barring a few subjects who had 'no reactions', they were confident in their ability to determine when a subject had missed the target. To predict the *intent* target, they tried to follow where the participant was aiming using eye gaze and body posture prior to the start of the throwing phase. In addition, they tended to assume that subjects would on average be inaccurate in the vertical axis rather than the horizontal axis. The heuristic is confirmed by the frequency data shown in Figure 4.

Table 5 lists the performance metrics averaged over both human labelers. Overall, they outperformed all models considered. Comparing the best case human labeler performance to our models, outcome prediction accuracy was 0.43 compared to the model accuracy of 0.34, congruence had a MCC score of 0.33 compared to model MCC of 0.29, and intent prediction accuracy of 0.33 versus 0.25 for the model (Section 6.2 discusses why MCC was considered the appropriate metric for intent-outcome congruence). We attribute this to the human ability to use the participant's body position prior to the throw, reason about the throwing motion employed, and extrapolate the ball's trajectory once it has left the camera's field of view. Human detection and interpretation of emotions and expressions is currently far superior to computational models. Finally, human ability to predict the ball's position even when it is blurry in the image is superior to the computational models.

# 6 Results

All models were evaluated using 5-fold nested cross validation. Training was carried out on a system running Ubuntu 18 equipped with a Ryzen 1700, Nvidia 2080S GPU, and 16GBs of RAM. All two-stream models were trained in Pytorch 1.5 with a batch size of 10, and the SGD optimizer with momentum of 0.9. The initial learning rates were 1.24e-2 for the RGB stream and 2.4e-4 for the flow stream. The learning rate was reduced to 10% of its value every 20 epochs. Training was stopped when the validation loss stagnated for 10 epochs to ensure convergence. The LSTM and CNN models were trained in Tensorflow 2.1 with a batch size of 5, Adam as the optimizer, and a learning rate of 2e-3. Training was stopped when the validation loss did not improve for 20 epochs. The mean value over all 5 folds is reported for all metrics unless otherwise specified.

| 0.46 | 0.10 | 0.09 | 0.07 | 0.23 | 0.38 | 0.28 | 0.05 | 0.09 |
| --- | --- | --- | --- | --- | --- | --- | --- | --- |
| 0.25 | 0.37 | 0.47 | 0.19 | 0.35 | 0.24 | 0.21 | 0.28 | 0.43 |
| 0.41 | 0.45 | 0.35 | 0.22 | 0.34 | 0.04 | 0.26 | 0.25 | 0.00 |

Fig. 7: Mean accuracy over the 5-folds for each target zone. Ordered left to right showing grids for the 0°, 45°, and 90° D435 camera views.

Table 6: Comparison of classification accuracy for outcome prediction using the D435 camera frames (mean/std-dev)

|  | 0° cam | 45° cam | 90° cam |
| --- | --- | --- | --- |
| LSTM (Ours) | 0.339/0.06 | 0.219/0.01 | 0.251/0.02 |
| Two-Stream (Li et al., 2020) | 0.245/0.04 | 0.179/0.03 | 0.178/0.03 |
| Human Labelers | 0.43/- | 0.23/- | 0.29/- |

Table 7: Intent-Outcome congruence MCC and F1 scores for the Pi camera frames (mean/std-dev)

|  |  | 0° cam | 45° cam | 90° cam |
| --- | --- | --- | --- | --- |
| 1D-CNN | MCC | 0.29/0.04 | 0.19/0.06 | 0.05/0.07 |
|  | F1 | 0.62/0.02 | 0.58/0.03 | 0.48/0.06 |
| Human Labelers | MCC | 0.14/- | 0.33/- | 0.23/- |
|  | F1 | 0.69/- | 0.74/- | 0.72/- |

## 6.1 Outcome Prediction

Only the 1035 throws that struck one of the nine target zones were used. To account for the differing number of throws per participant, the throws were split on a per participant basis using 80% for the training set and 20% for the testing set. This ensured the number of throws per participant in the split sets were representative of the dataset. In addition, the splits were stratified using the *outcome* target to better represent the distribution in the dataset. For a perfectly balanced split, we would expect a mean of 0.11 and a std-dev of 0.0 for the percentage of samples belonging to each target zone that makes up the set. For the 5 splits used, the overall mean was 0.108 with an average std-dev of 0.019 which meant our dataset was balanced and had a similar number of samples for each *outcome* target. The training set was further split into a training and validation set using a 80:20 ratio to determine when to stop training the model. Using the testing set for this purpose can lead to over-fitting on the test set and an inflated accuracy metric.

Table 6 compares the model introduced in Section 4.3 to the Two-Stream scheme (Li et al., 2020), and the human labeler baseline. Predicting the majority class for the 5-folds would result in a mean accuracy of 0.14 (std-dev 0.005). The performance of Li et al.'s Two-Stream approach is discussed in detail in Section 7 and in summary our model improves performance by 33% over all 3 views compared to the Two-Stream approach. Our model was able to get closer to the human labeler accuracy in the 45° and 90° cases since these views made it hard for the labelers to perceive depth and use their knowledge of throwing actions to extrapolate ball trajectory. Figure 7 shows the proportion of test cases that were correctly predicted for each target zone. This illustrates that our model is able to differentiate ball trajectories in the X and Y axis and the accuracy is not a result of skewed performance on few target zones. The poor performance on target zone 9 can be attributed in part to the 45° and 90° camera being increasingly perpendicular to the ball trajectory which worsens depth determination from RGB camera images which are 2D projections of the 3D trajectory of the ball.

Predicting the outcome of the throwing task using only RGB data of the throwing motion is a very difficult problem. Inter-subject variation in the form of varying throwing actions and limb lengths, and intra-subject variation in the form of changing ball release speed, release point, and the use of multiple overhand throwing actions greatly increases the complexity of this problem. These variations, caused by differing body geometry, results in the inability to accurately predict the ball's trajectory using body pose as current computer vision techniques lack the ability to precisely and consistently detect joint position with sub-millimeter accuracy in the absence of tracking cameras. Moreover, the same throwing action can result in different ball trajectories if the release point is minimally altered and it is not possible to accurately track the hand to determine the release point with our current setup.

6.2 Intent-Outcome Congruence Prediction

In this case, the 5-folds for cross validation were created similar to outcome prediction using only the 1035 throws that struck one of the nine target zones. Throws from each participant were split using 80% for the training set and 20% for the testing set. Unlike outcome prediction, the splits were stratified using the dummy variable, congruence, defined as,

$$congruence = \begin{cases} 1, outcome = intent \\ 0, outcome \neq intent \end{cases} \quad (2)$$

This led to imbalanced folds where on average 63% of the cases belonged to the positive class in this binary classification problem. The training set was further split into a training and validation set using a 80:20 ratio.

When training our neural network, the loss function was biased using class weights calculated using the distribution of the training set to account for the imbalanced training set. When a binary classification dataset is imbalanced, the F1 score which computes the harmonic mean of the recall and precision can be used to evaluate the model. While we provide the F1 scores, we also discuss the Matthews Correlation Coefficient (MCC), also referred to as the phi coefficient, as it can be shown to be superior to the F1 score (Chicco and Jurman, 2020). In addition, the F1 score is not normalized and the score is not symmetric when the classes are swapped. MCC is easier to interpret with a score of -1 representing an inverse prediction, 0 an average random prediction, and +1 a perfect prediction. To illustrate the utility of the MCC score, consider that for a random classifier on this task, the mean MCC over all folds was 0.02 (std-dev 0.05) and the F1 score was 0.48 (std-dev 0.03). For a classifier that predicted only the negative class the MCC was 0 and the F1 score was 0.27 whereas for one that predicted only the positive class the MCC was 0 and the F1 score was 0.38. The MCC score correctly captures that these models are not predictive which is very hard to glean from the F1 scores. For the sake of completeness, we present the F1 scores in addition to the MCC.

Table 7 shows the metrics for the intent-outcome congruence model discussed in Section 4.4 compared against the human labelers. We do not compare to Li et al. as their work assumes that the *intent* target is always coincident with the *outcome* target and only perform what we refer to as outcome prediction.

Both the labelers and our model outperform a random average model which had an MCC of 0.02 and F1 score of 0.48, and a majority class predictor with MCC of 0 and F1 of 0.38. This provides evidence that the model is able to predict the congruence at a rate above chance and that the subject's internal state change in response to the outcome of their action is expressed through facial emotions and can be captured using RGB cameras. Intent-Outcome congruence prediction is one where we expect humans to vastly outperform the algorithm. Surprisingly, our algorithm outperforms the human labelers in the front-on 0° camera view with an MCC of 0.29 compared to the humans at 0.14.

Our model performance degrades as we get more side-on and the camera moves away from capturing frontal videos of the subject. This degradation can be attributed to poor facial emotion recognition when the subject's face is not captured head-on as the pre-trained emotion model and available facial emotion datasets are based on close-up frontal images of the face. Another major issue was that taller participants tended to move outside the field of view of the 45° and 90° cameras during their throwing action even though they were instructed to minimize translation towards the target. These two issues do not affect the 0° camera which captures front-on videos of the subject.

Table 8: Effect of probability matrix on intent prediction accuracy (mean/std-dev)

| Dataset Avg. | Subject Specific | Uniform |
|---|---|---|
| 0.451/0.05 | 0.374/0.05 | 0.113/0.03 |

Table 9: End-to-End accuracy for the three camera views (mean/std-dev)

|  | 0° cam | 45° cam | 90° cam |
|---|---|---|---|
| Ours | 0.253/0.06 | 0.157/0.01 | 0.180/0.01 |
| Labelers | 0.33/- | 0.16/- | 0.25/- |

6.3 Intent Prediction

For this experiment, the 80:20 train/test splits were stratified using the *intent* target and only trials where the subject hit one of the nine target zones were used. The resulting splits were balanced as the data collection procedure ensured subjects aimed at all nine target zones equally. Subjects were also equally likely to make a mistake regardless of target zone, hence ignoring trials where they completely missed the target did not introduce any significant imbalance in the splits. We therefore use accuracy as the evaluation metric.

Here we evaluate the performance of intent prediction using the probability matrix as described in Section 4.4.2 for only those data samples where incongruence is detected i.e. the *outcome* is not the *intent*. This allows us to evaluate performance in isolation of the outcome prediction pipeline as when incongruence is not detected, the intent prediction accuracy is directly tied to outcome prediction accuracy.

Three different probability distributions were considered: A uniform probability distribution used as a baseline, probabilities based on all trials in the training set, and specific probabilities generated for each subject calculated using only the trials they completed. Table 8 shows that using the entire dataset to generate the probabilities outperforms only considering subject specific trials. Both these methods perform far above random or using a uniform distribution. The low number of trials some subjects completed in the hour of data collection explains the poor performance of the subject specific model. With more data, we expect the accuracy of a subject specific model to improve. Humans tend to have biases in their movement patterns which would be reflected in hitting certain target zones more frequently. In the case of our throwing dataset, the higher accuracy of the model which uses the entire dataset suggests that the subjects had similar patterns of mistakes.

6.4 End-To-End Pipeline

Finally, we evaluate the performance of intent prediction on both samples that contain and do not contain mistakes in contrast to the previous section. That is, we evaluate the end-to-end system which uses the predicted outcome and intent-outcome congruence prediction to determine subject intent. For this experiment, we have to ensure that the same training/test split is used to train all the models to prevent data leakage. The training splits were stratified using the *intent* targets since this best represents the throwing task. The overall dataset probabilities were used for intent prediction. For the end-to-end system, the best accuracy was found to be 0.253 (std-dev 0.06) when the 0° camera was used.

Table 9 shows the end-to-end prediction accuracy for the various camera views with the D435 camera data used for outcome prediction and the Rasberry Pi camera data used for intent prediction. We found that the highest end-to-end intent prediction accuracy was obtained using the 0° camera view which corresponds to the view with the highest outcome prediction accuracy discussed in Section 6.1. This highlights the high degree to which the outcome prediction accuracy affects the accuracy of the end-to-end system, and when the outcome is predicted incorrectly, this has a cascading effect on intent prediction. The model outperforms a random predictor which has an accuracy of 0.11 in all cases. But, compared to the human labelers, we were only able to match their performance in the 45° view. This can be attributed to the superior labeler outcome prediction informed by learning outside of this task allowing them to infer ball trajectory from the videos better than the proposed outcome prediction model.

7 Comparison to Closely Related Work

In this section we address the inability to replicate the model performance of the Two-Stream architecture reported by Li et al. and discuss,

- Choice of train/test splitting (Li et al., 2018).
- Use of the test set to select the best model parameters during training (Li et al., 2020)

– Differences in the dataset used in Li et al. (2020) and our throwing dataset

(A) Li et al. (2018) and (B) Li et al. (2020) use a 9 target experimental setup with a single camera pointed towards the subject which corresponds to the 0° camera in our version of the experiment. (A) explores data augmentation methods for improving performance while, (B) proposes the use of the two-stream deep learning architecture. In (A) a single human participant was used to create a dataset consisting of 256 training trials and 36 testing trials. In (B) a larger dataset was created using 6 human subjects each performing 90 throws to obtain 432 training samples and 108 testing samples, and introduces the two-stream architecture as an improved model over the CNN used in (A). Both (A) and (B) report an outcome prediction accuracy in the range of 50%-75% for the data augmentation and sampling techniques they developed respectively. Our evaluation of the two-stream architecture on our dataset introduced in Section 3.2 resulted in much lower performance metrics listed in Table 6.

Inspection of the code used for (A) available from the project's publicly available Github (Zhang, 2018) revealed an issue with the strategy used to create the train and test sets resulting in data leakage between these sets. In (A) the frames from all the throwing trial videos were extracted and collated to form a large dataset of images which was then shuffled and split into a train, validation, and test set. While this is a common way of creating splits for image recognition tasks, it is not appropriate for video classification as the training and testing sets would contain frames from the same video. To avoid data leaks, the videos must first be assigned to either the train or test set before extraction of the frames.

The code and data used for study (B), which is currently not publicly available, was obtained by contacting the authors. There was no data leak in (B) and the train/test sets were split appropriately on a per video basis. The code we received evaluated the model on the test set after each training epoch, saving the model with the highest accuracy on the test during the training phase. This overfits the model to the test set. A validation set should have been employed to avoid this issue and this method of parameter selection would severely impact the model's ability to generalize to unseen data. From the code it was unclear if K-Fold cross validation was employed. Training the Two-Stream model using the code sent to us by the authors on the dataset used in (B) resulted in performance shown in Table 10 which is lower than that reported in (Li et al., 2020). This variation in performance can be attributed to the different train/test splits we generated and random initialization of the network parameters among other factors.

We conclude by evaluating the performance of our outcome prediction method on the dataset used in (B). We used 5 fold cross validation. For fair comparison, we also used the test set to select optimal model parameters. Table 10 shows that our method is able to match the accuracy of the two-stream model.

7.1 Difference between our dataset and (B) Li et al. (2020)

The most striking difference is the speed at which the participants in (B) throw the ball. In our dataset participants threw the ball significantly faster as the only limitation placed on them was the use of an overhand throwing action. This resulted in many frames in our dataset containing a blurry ball and arm captures due to the joint and ball velocities even though both datasets captured images at 30FPS.

Another difference is the number of frames for which the ball was visible after release. On average the ball was visible for 4-6 frames after release in our dataset compared to 10-12 frames in (B). The Kinect V2 they used has a field of view (FoV) of 84.1x53.8 compared to the D435 with a FoV of 64x41. These factors made it easier to determine the outcome target based on the location of the ball in the camera frame in (B). Since their dataset is private, we include a graphical representation of the location of the ball and subject in the final frame before it leaves the camera field of view for the 9 target zones in Figure 8. Compared to the location of the ball in our dataset, shown in Figure 9, where the ball moves towards the top of the image, it is far easier to determine the *outcome* target in the version of the dataset used in (Li et al., 2020) based on ball position and size in pixels. Thus the Two-Stream architecture which uses majority voting of the individual frame predictions to predict the outcome target performs poorly as the location of the ball in the frame corresponding to each target zone becomes harder to separate. An LSTM on the other hand with input features containing the location of the ball is able to use temporal information of the ball's trajectory to predict the outcome with higher accuracy.

8 Conclusion

Physically dynamic tasks introduce unique challenges from the perspective of intent recognition due to the increased number of unforced errors humans make while performing them. In our throwing dataset out of a total of 1227 throws, 582 were mistakes with the subjects completely missing the entire target grid in 192 of these mistakes. Prior work that considered the throwing task ignored throws that resulted in mistakes introducing a one-to-one correspondence between the subject's intent and the outcome. In realistic situations, mistakes must be accounted for and the outcome cannot be considered to be the human's intent. We propose a method that leverages the surface cue of facial expressions allowing us to predict when the subject believes they have committed a mistake. Using this information we inform our prediction of intent in the context of the throwing task.

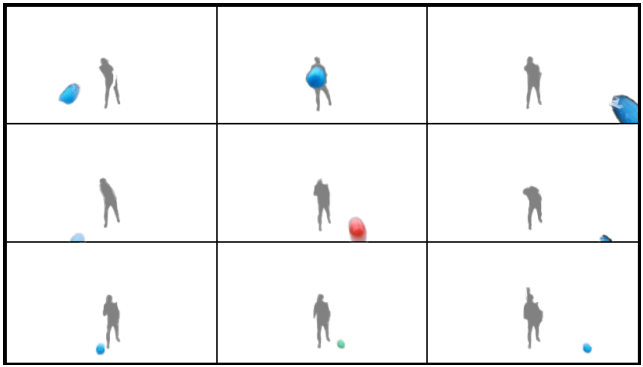

Fig. 8: Anonymized image showing the position of the ball in (Li et al., 2020) for each of the 9 target zones when thrown by a single participant. Even with the naked eye one can differentiate between the outcome targets especially which column of the target grid the ball might strike. Image at the top left represents target zone 1, while the image at the bottom right shows target zone 9.

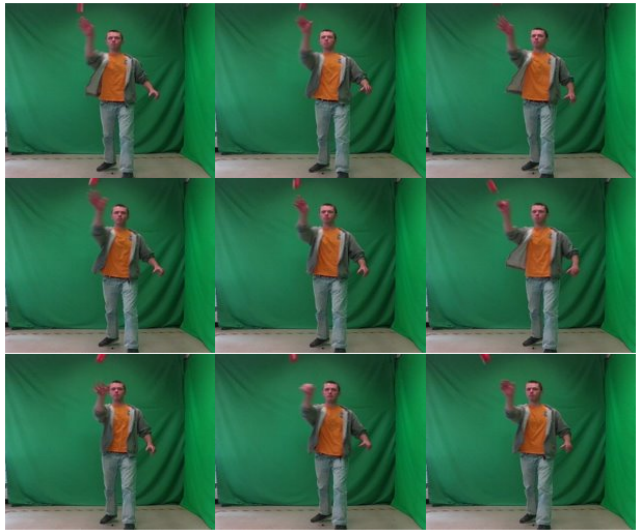

Fig. 9: Position of the ball for each of the 9 target zones for a single participant in our dataset showing the difficulty of determining the outcome target from the ball position in the frame. Image at the top left represents target zone 1, while the image at the bottom right shows target zone 9.

Table 10: Accuracy of two-stream model introduced in (Li et al., 2020) compared to our method presented in Section 4.3. Shows our model is able to match performance on the dataset used by Li et al. (2020). Results shown for all 5-folds of cross validation.

|  | 1 | 2 | 3 | 4 | 5 |
|---|---|---|---|---|---|
| RGB Stream | 0.41 | 0.50 | 0.30 | 0.34 | 0.50 |
| Flow Stream | 0.38 | 0.33 | 0.27 | 0.32 | 0.50 |
| Avg. Fusion | 0.40 | 0.50 | 0.29 | 0.34 | 0.51 |
| Our Method | 0.45 | 0.43 | 0.49 | 0.36 | 0.47 |

We propose the use of an LSTM to predict the outcome from the partial ball trajectory and show that it outperforms the majority voting prediction used by the Two-Stream architecture previously used for this task. Our 1-D CNN which uses features learnt from a facial emotion dataset is able to outperform human labeler performance in predicting the presence of a mistake when given front-on videos. Using the output of these two models along with priors learnt from the frequency of mistakes, we provide an end-to-end system that takes as input a video of a throw and predicts the *outcome* and *intent* for a throwing task consisting of 1035 throws generated by 10 people.

While our end-to-end pipeline performs better than chance and matches human performance on the 45° camera view, it is not able to outperform the human labelers. The use of an emotion dataset with front-on facial images results in degraded performance as in the 45° and 90° views which does not affect the labelers. Although we have improved outcome prediction, vision based approaches under-perform when compared to motion capture systems. The main culprit is the resolution and precision by which both these methods estimate human joint position. The advantage of vision-based approaches is that they can be used on robots in the field unlike MOCAP systems that require expensive equipment and are constrained to a calibrated environment. We also show that a two-stream architecture for throwing tasks is ill-advised unless the location of the ball in the frames is well seperated for each target zone. When the position of the ball is not well differentiated, we propose the use of an LSTM that is able to leverage temporal information and extrapolate the trajectory of the ball to predict the target zone.

Since we use a subject's reaction to the observed outcome to predict if a mistake is likely, it is not possible to predict the intent earlier than 30 frames after the completion of the task. We envision this pipeline to be utilized in applications where a robotic teammate would assist the human when it has detected that the outcome observed was not the intent of the human. In such tasks, prediction of the intent is only necessary before the next step of the task has to commence. For example, in our throwing task the robotic teammate might help by striking the correct target if the human missed the target. As long as there is about 1 second in addition to the time taken by the robot to perform the actions available to complete the task, this late prediction should be acceptable for a particular application. On the other extreme, if the task involves predicting the intent before the ball thrown by the human has hit the target, or before the human has reacted to the outcome, then our method which relies on capturing the reaction of the human would not be suitable.

Advancements in the prediction of temporally stable 3D poses from RGB-D data would help reduce joint position errors and improve performance of outcome prediction using 3D pose as input compared to 2D pose. Unfortunately due to the global COVID-19 pandemic we were extremely limited in our ability to conduct human subject research. Future work could test these ideas using additional human subjects with varied backgrounds and limb lengths as the type of cues exhibited could differ significantly. The advantage of using emotion cues is that these cues are largely agnostic to a person's physical characteristics, but the same cannot be said for outcome prediction using human pose. Normalisation of lengths based on a reference length might overcome this but it is unclear if different body types inherently bias the types of surface cues exhibited. We must also consider that emotion cues likely vary with one's culture, age, and gender, and this information is rarely available for the emotion datasets we transfer learn from.

We argue that changes to the task setup such as orientation of the target and number of targets would be unlikely to change the validity of the method developed as it is probable that subjects exhibit comparable reactions to the outcome of a throw. Even if the task type was changed to require a different physical action, for example rolling a ball towards targets, or pushing an object towards multiple targets, we believe our method would be able to predict outcome-intent congruence as long as the task meets the key criteria of being one where mistakes are likely. This arises from the hypothesis that the surface cues of facial expression and body gestures arises from the subject's reaction to an uncertain outcome when performing a skill based action. It is possible that stronger cues are generated if an element of chance/luck in introduced to increases the uncertainty of the outcome. Inversely, when the outcome is certain, for example the task of placing the ball in one of many target boxes without time pressure, the intensity of the surface cues exhibited, if any at all, would be very low making them undetectable.

The adaptability of this technique is illustrated by the use of facial emotions as a feature to detect incongruence in a task that did not have dynamic motions but introduced uncertainty in the form of incomplete information. Surendran et al. (2021) showed that in a card game where the human intends to deceive the robot, there are patterns in the sequences of the perceived surface cues that can be used to detect incongruence which was then used to predict the intent to deceive on par with human performance on the task.

Future work should also consider variations that affect the difficulty of the task. Varying the distance to the target and sizing of the target grid can affect not only subject performance but model prediction accuracy. The order of the target zones aimed at were randomly selected to minimize increase in accuracy due to practice/learning effects. Participant accuracy would also be negatively affected by fatigue as time goes on. We did not attempt to quantify the effects of practice and fatigue on accuracy in our throwing task as this would require a longitudinal study with repeated data collection on the same subject to evaluate improvement in performance. A longitudinal study deriving from this research is another opportunity for future work as this would inform how an intent recognition system can adapt to a particular subject over time. Finally, fusion of images from multiple viewpoints to generate prediction could also improve performance by relying on the advantages of each view. This initial attempt at predicting intent using pose and emotion cues shows promise and at the very least provides a method that can help detect mistakes and the lack of congruence between the observed outcome and the expected outcome.

**Conflict of Interest**

The authors declare that they have no affiliations with or involvement in any organization or entity with any conflict of interest regarding the subject matter or materials discussed in this manuscript.